# Toolbox Spotter: A Computer Vision System for Real World Situational Awareness in Heavy Industries

**Stuart Eiffert, Alexander Wendel, Peter Colborne-Veel,
Nicholas Leong, John Gardenier, and Nathan Kirchner**

**Presien**

{stuart.eiffert, alexander.wendel, peter.colborne-veel, nicholas.leong, john.gardenier, nathan}@presien.com

**Abstract -**

The majority of fatalities and traumatic injuries in heavy industries involve mobile plant and vehicles, often resulting from a lapse of attention or communication. Existing approaches to hazard identification include the use of human spotters, passive reversing cameras, non-differentiating proximity sensors and tag based systems. These approaches either suffer from problems of worker attention or require the use of additional devices on all workers and obstacles. Whilst computer vision detection systems have previously been deployed in structured applications such as manufacturing and on-road vehicles, there does not yet exist a robust and portable solution for use in unstructured environments like construction that effectively communicates risks to relevant workers. To address these limitations, our solution, the Toolbox Spotter (TBS), acts to improve worker safety and reduce preventable incidents by employing an embedded robotic perception and distributed HMI alert system to augment both detection and communication of hazards in safety critical environments. In this paper we outline the TBS safety system and evaluate it's performance based on data from real world implementations, demonstrating the suitability of the Toolbox Spotter for applications in heavy industries.

## 1 Introduction

In 2018 there were 99 fatalities in the heavy industries of Transportation, Agriculture, and Construction alone in Australia, accounting for 69% of all workplace fatalities across all industries [1]. The vast majority of these (71%) were directly related to vehicle collisions and impacts with other moving machinery.

ToolBox Spotter (TBS) is an embedded robotic perception and distributed alert system for use in heavy industries that works to supplement existing safety procedures in these critical environments. It addresses both the detection of hazards that may result in a collision or injury, as well as the effective communication of these hazards to vehicle operators and nearby pedestrians. The TBS is a modular system consisting of local sensor nodes, a central processing node (CPN), and distributed alert devices. This forms an intelligent detection system and alert network for use on vehicles, mobile plant and machinery, and on infrastructure as a safety control measure.

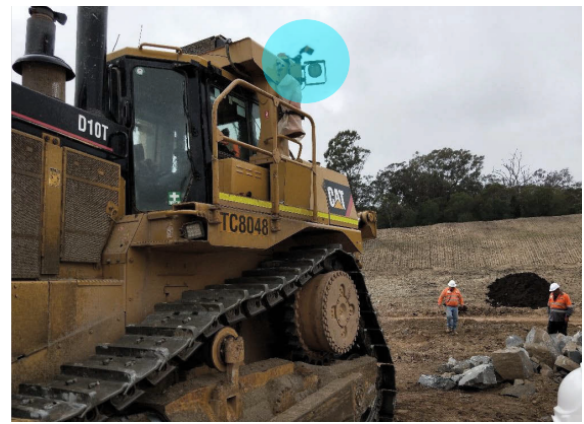

Figure 1. *The Toolbox Spotter (TBS) hazard awareness system in use on heavy machinery at a construction site. The sensor node (highlighted in blue) is alerting the operator to the presence of the two people standing in the vehicle's blind spot.*

Existing approaches to safety include the use of a hierarchy of controls which aim to eliminate and mitigate risks where possible through the use of standard policies such as task isolation and the use of personal protective equipment (PPE). Whilst these procedures can be effective when correctly employed, it is not always practical to completely remove a risk during normal operations. Additionally, worker distractions and lapses of attention can greatly reduce the effectiveness of these approaches. Recently, proximity based collision detection devices combined with wearable RFID tags have been used in safety systems in construction [2]. However, these systems cannot always differentiate between objects when RFID tags are not worn and crucially rely on human behaviour and supervision to ensure this.

In this work we describe our solution to the problem of safety around moving vehicles and plant in heavy industries, the ToolBox Spotter. Section 3.1 provides an overview of the components of the TBS system.

We also evaluate our proposed system in Section 5 on three diverse datasets of real world video clips, chosen to reflect current implementations of the TBS on construction sites. We outline the performance of the system as well as addressing usage of the system with consideration of its application as a human in the loop safety system.

## 2 Background

### 2.1 Safety in Heavy Industries

A comprehensive study of occupational safety in the construction industry [3] detailed that the primary conditions influencing safety performance in a workplace were not just the organisational procedures and policies in place, but the individual attitudes towards safety, including personal engagement, the taking of responsibility, and prioritisation of safety. Whilst the proper use of control procedures are critical in the elimination of avoidable risks, there will always be situations in which a degree of inherent risk is unavoidable. In these cases there is a clear need for safety controls that are not dependent on individual behaviours, such as the use of PPE.

Previous approaches to reducing vehicle collisions in heavy industry have made use of proximity sensors, including RADAR and ultrasonic based systems [4, 5]. These devices have less capability of differentiating between different types of objects, and so have been used alongside body tags, using RFID or magnetic fields [2]. Whilst these systems can work effectively to reduce the risk of collisions they again depend on individual behaviours, requiring each worker to wear the tag as additional PPE. These signals are also significantly impacted by conductive materials and can be obstructed by nearby vehicles and human bodies [6].

### 2.2 Computer Vision based Safety Systems

The use of computer vision object detection in safety systems is well established in areas including automotives and manufacturing, where these systems are integrated into the vehicles or fixed infrastructure [7, 8]. Similar systems have also seen recent application in mining vehicles, where they are used for both personnel and vehicle detection in less structured environments [9]. These systems have been shown to effectively detect hazards in heavy industries, but are generally used within a completely automated process, rather than forming a human in the loop system that communicates these hazards to relevant workers. Additionally, these safety systems are often built in to the hardware in which they are used, minimising their capability to be retro-fitted to existing vehicles and infrastructure or be used in a portable manner. Other applications include the use of computer vision technologies for safety management, in which the detection of workers' locations, activities and behaviours from surveillance cameras is used to inform management strategies for mitigation of future risks [10]. Whilst these approaches can help decrease incidents over a longer time period, they do not communicate immediate hazards to vehicle operators or workers.

Human in the loop safety systems require a thorough understanding of the interaction between humans and automation [11]. The system's behaviour should be intuitive to the human and not cause extra cognitive load. An example of intuitive behaviour is to only alert an operator of the presence of a distant person when the person is approaching the operator, not when the person is moving away. This example only holds for distant objects, as all close objects should result in an alert to the operator. Behaviour of a human operator will be influenced to be more positive and safe in a well-designed human in the loop safety system.

Our TBS system can communicate with human operators via haptic, visual, and audio alerts. Additionally, a halo-light can be placed on top of a vehicle to alert people approaching the vehicle to the detection of their presence. This acts to enhance 'positive communications', a procedure where a person walking behind or alongside a construction vehicle must establish that the operator is aware of their presence.

## 3 System Details

### 3.1 Overview

The TBS is composed of a network of connected devices, including a single Central Processing Node (CPN), multiple camera sensing nodes, and a Human Machine Interface (HMI) consisting of distributed AlertWear alert devices and a user interface (UI) tablet device. Each sensing node passes 2D images to the CPN where the Alert Pipeline, outlined in Figure 2, is used to determine when to send an alert to the distributed AlertWear devices. The behaviour of the Alert Pipeline is configurable by the operator using the UI device, allowing the setting of which types of hazards to be alerted to, including people, light vehicles, heavy vehicles, and demarcations such as traffic cones and bollards. Additionally, the operator is able to configure exclusion zones within each sensing node's field of view, in order to define alerts to a region of interest.

### 3.2 Alert Pipeline

The main components of the alert pipeline are outlined in Figure 2. Input images are first preprocessed, including a check of image quality used to advise operators if the quality is outside operating conditions via the UI device.

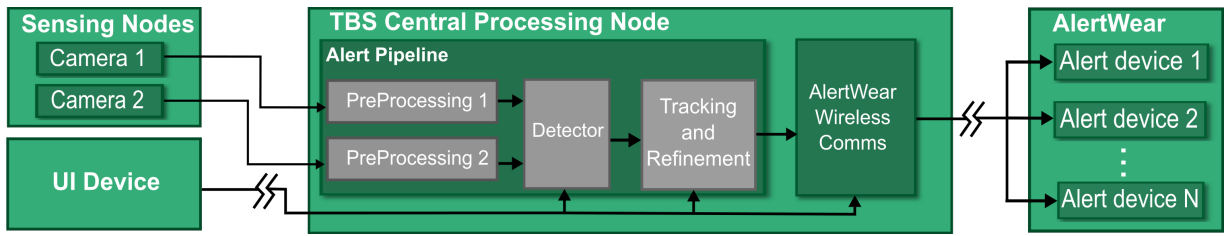

Figure 2. *Network architecture of the TBS system, illustrating communication between sensing nodes, central processing node (CPN), UI device, and distributed AlertWear devices. Alert Pipeline is detailed in Section 3.2*

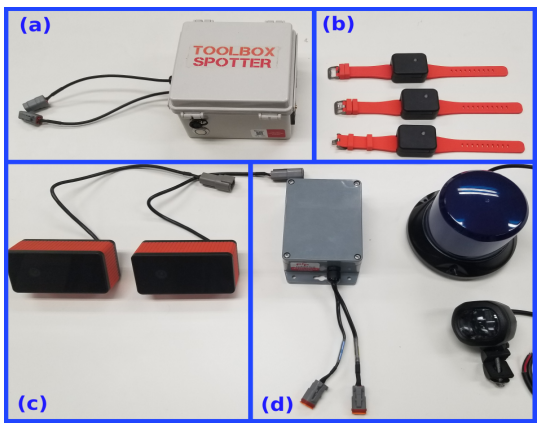

Figure 3. *The physical TBS system, showing the Central Processing Node (CPN) (a), Alertbands (b), Sensing Nodes (c) and additional AlertWear devices (d) including the Alertbeacon (top right), halo-light (bottom right) and expansion node (left) which allows connection to external devices.*

Object detection in the camera frame is performed using convolutional neural networks (CNNs) [12, 13, 14], which have been pretrained as per Section 3.3. In order to minimise duplicates and false detections, improve detection localisation accuracy, smooth alerts, and to determine the final confidence of each detection, the output is further processed before alerts are communicated to the operator. Objects are tracked between frames and if an object's confidence exceeds the user defined threshold, an alert is issued to each connected AlertWear device. Tracked objects are finally filtered based on selected class types in the UI, and by region of interest, if an exclusion zone is being used (see Section 3.4).

### 3.3 Application Specific Training

We make use of a proprietary dataset of 15870 labelled construction specific scenes to fine-tune our models after pre-training on publicly available datasets, e.g. [15]. To evaluate detection performance, the dataset is split randomly into a train and a test set. Dataset balance and labelling quality is critical to the trained model performance. Data augmentation is performed during training, allowing the model to generalise better to new unseen data. When our system is implemented in new environments, images from the new environment are added to the train and test dataset. Semi-automatic human-in-the-loop labelling is performed using the existing model to provide labels.

### 3.4 HMI

**AlertWear**
The purpose of the AlertWear system is to communicate detected hazards and risks to all relevant workers in a clear and non-intrusive manner. This system involves a wireless mesh network (IEEE 802.15.4 std) of devices which can expand to accommodate user selected devices.

The main means of communication is an Alertband, worn by workers which vibrates to communicate various alerts. This interface was chosen based on initial pilot studies conducted of the system on a major construction company's operational sites, in which users deemed visual and audible warnings either too distracting or not noticeable enough, leading to the misuse and non-use of the system. From these studies it was found that a two second pulsed vibration was the most effective means of communicating an alert. The AlertWear system also supports visual and audible warnings through the use of the halo-light, which can be mounted on vehicle to illuminate exclusion areas, and the Alertbeacon which can provide area wide alerts, as shown in Figure 3.

**Operator Interface**
A UI app has been developed to allow configuration of each device, including checking sensor field of view, selection of detection classes, setting of any exclusion zone and the changing of Alert Pipeline settings. Alert Pipeline settings are input by using sliders which allow configuration of the detection and tracking thresholds. Changing these values will alter the false positive and false negative rate.

In this work we have tested three different alert modes based on different configurations of these sliders: (1) *Default*; (2) *Reactive*; and (3) *Certain*, where *Reactive* aims to minimise alert delay, *Certain* minimises false positives, and *Default* aims to strike a balance between the two other modes. This UI also allows the setting of user defined exclusion zone, which can be used to limit alerts to detections that intersect with an area of interest in the camera frame. An example of an exclusion zone is shown in Figure 5 (c).

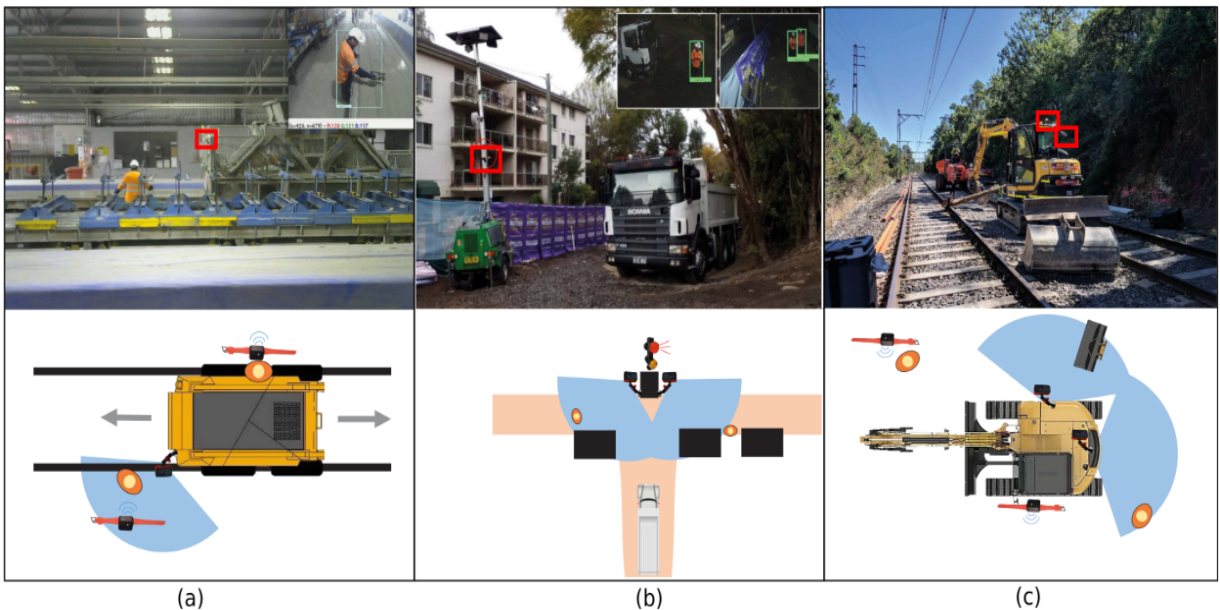

Figure 4. *Real world use cases of the TBS are shown in the top row, with the location of sensing nodes highlighted in red and camera views as inset if available. The bottom row is a top down illustration of the setup, showing the location of sensing nodes, AlertWear, and objects, as well as the field of view of each sensing node in blue, and people as orange ellipses . Example (a) demonstrates a single sensing node on a manufacturing line covering a blind spot of the mobile machinery from the operator's viewpoint. (b) shows two nodes and an Alertbeacon installed on infrastructure at intersection alerting oncoming traffic to the presence of hazards around a blind corner. (c) shows two nodes covering a rail vehicle's blind spots.*

## 4 Empirical Evaluation

Evaluation of the overall TBS system has been conducted in order to determine performance across a variety of real world scenarios and visual variations. This evaluation has been done specifically with the class type of people, rather than other hazards and vehicles in the images for the sake of clarity and to to provide an indicative example of performance on the highest priority class.

### 4.1 Datasets

The TBS is currently implemented across a number of real world heavy industry sites, including construction, mining, agriculture and manufacturing. Within these implementations the TBS is being used on vehicle, on infrastructure or buildings, and within manufacturing lines.

Figure 4 provides examples from each implementation, showing both the system in use and a top down diagram of installation. The datasets used in this work have been chosen to reflect these real world use-cases, and include:

1. **Vehicle**: On vehicles in an off-road environment
2. **Indoor**: On mobile machinery for indoor use.
3. **Infra**: On infrastructure in a road environment

The Vehicle dataset contains a total of 17 clips, Indoor dataset 10 clips, and Infra dataset 7 clips. Each clip contains approximately 20 seconds of video. Example images from each dataset are shown in Figure 5. These clips cover the following range of visual variations, encountered during real world implementations of the TBS:

- Overexposure and glare
- Varying target obstruction
- Image degradation (blur and dust)
- Varying target distance
- Varying clutter in image

Additionally, a second subset of clips has been created for supplementary testing, which we refer to as 'Outlier Clips'. This subset was compiled based on a subjective measure of how difficult it was for a human labeller to initially identify the person in the clip and includes examples

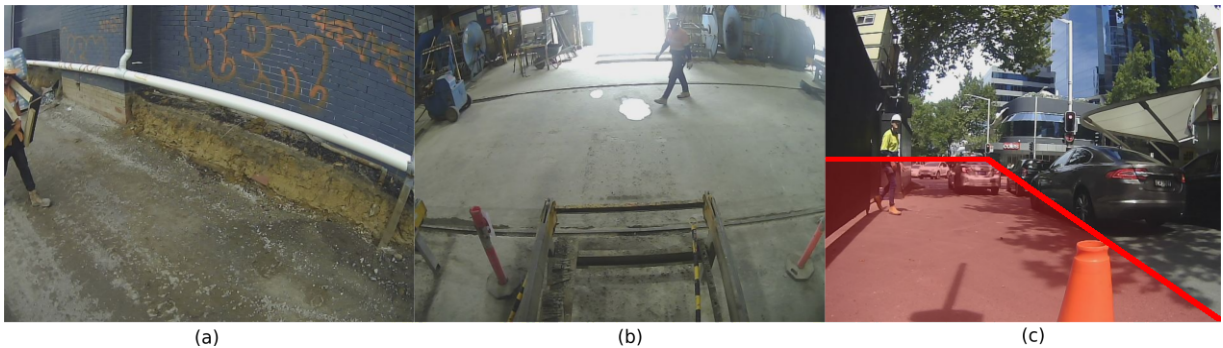

Figure 5. *Example TBS viewpoint images from each of the test datasets as outlined in Section 4.1, illustrating the visual variance tested. (a) is mounted on a vehicle in an outdoor environment, (b) on moving machinery indoors, and (c) on stationary infrastructure on road. Example (c) illustrates how a user defined exclusion zone (shown in red) is displayed in the UI, as described in Section 3.4.*

that the average user would not be expected to see within a reasonable length of time.

These examples are a result of:

- The sensing node being set up without a clear view of the area of interest
- The person of interest being very heavily occluded or located very far from the sensor
- The person of interest appearing in an unexpected area of the frame
- Extreme visual aberrations including glare or usage at night

Testing on the 'Outlier Clips' subset is detailed further in Section 6. Figure 7 provides two typical examples of images from this data subset.

### 4.2 Metrics

Performance of the TBS system has been measured on the following metrics for all datasets:

- **Precision** - True positives over all detections
- **Recall** - True positives over ground truth occurrences
- **Alert %** - Proportion of people resulting in an alert
- **Alert Delay** - Time from first appearance to initial alert for each person

Precision and recall have both been calculated on a frame-wise basis. Alert % is calculated based on the number of completely missed alerts. For instance, a clip containing a single person walking through the scene would score 100% if an alert was sent whilst the person was in frame, or 0% otherwise. Alert Delay is calculated based on the time between a person entering the frame and the an alert being sent. Testing has been repeated for each of the three detection modes outlined in Section 3.4 determined by the system configurable parameters, including: (1) *Default*; (2) *Reactive*; and (3) *Certain*. Testing has then been repeated 5 times for each dataset.

### 4.3 Implementation

During testing each recorded dataset has been played by an external host computer over an ethernet network. This has been done using the ROS framework, duplicating the logged video over two separate streams to replicate usage of two sensing nodes. The output of the TBS has then been taken prior to the TBS CPN passing the alerts to the AlertWear wireless comms module. Alerts were then streamed back to the host using the same network.

Measured round trip network latency between host computer and TBS (83.0ms) has been deducted from all alert delay calculations and replaced with the measured real world sensing latency of the used cameras (67.0ms). Additionally, the time taken to send an alert over the AlertWear network has not been included in measures of alert delay. As this network uses IEEE 802.15.4 standard, we instead refer to previous testing which has shown a round-trip latency of 18ms for a 'single hop' less than 100m line of sight and 100ms for 4 hops. [16].

## 5 Results

Both *Default* mode and *Reactive* mode were able to correctly detect all people present in each clip, whilst *Certain* mode achieved an average accuracy of only 96.67%. The *Reactive* mode was also able to increase recall compared to *Default*, meaning that more frames containing a person were correctly identified as such. However this came at a cost to precision, meaning that false positives grew significantly. Conversely, *Certain* mode was able to increase precision at a cost to recall.

Figure 6 illustrates the distribution of delays for each tested mode across all datasets. Both *Default* and *Reactive* modes achieve significantly less delay than *Certain* mode, with peaks of 200ms compared to 1000ms. This equates to over 50% of all alerts occurring with a delay of less than 600ms for *Default* and *Reactive*, and 1000ms for *Certain*. This delayed response is due to the increased requirement that *Certain* mode has with regards to detection confidence. An alert will only be sent when the system has gained increased confidence in the likelihood that there is a person present. Whilst this does lead to slower reaction times, it also greatly reduces the number of unnecessary alerts, as reflected by the increased precision in *Certain* mode. It should be noted that the Infra dataset yielded a large number of delayed alerts for all modes. As the Infra dataset is significantly smaller than either the Vehicle or Indoor datasets (see Section 4.1), it is likely that some significantly harder clips have a greater detrimental influence on the resulting metrics. This can also be seen to a lesser extent in the Vehicle and Infra datasets in which minor second peaks occurs.

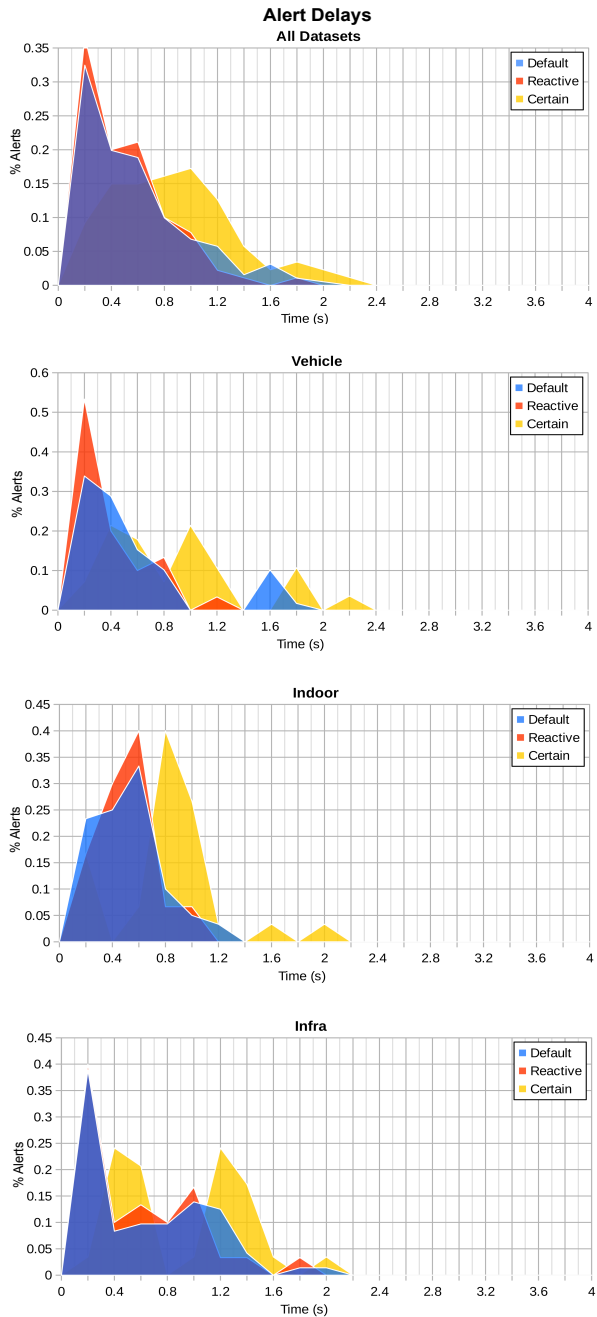

Figure 6. *Delays for each tested mode across all datasets. Delay refers to time between an object becoming present in the camera frame until the time at which an alert is passed to the TBS AlertWear. Combined result for all datasets is shown on top.*

| Dataset | | Mode | | |
|---|---|---|---|---|
| | | *Default* | *Reactive* | *Certain* |
| **Precision** | *Vehicle* | 0.845 | 0.826 | **0.891** |
| | *Infra.* | 0.751 | 0.547 | **0.863** |
| | *Indoor* | 0.925 | 0.868 | **0.955** |
| | *AVG* | 0.841 | 0.747 | **0.903** |
| **Recall** | *Vehicle* | 0.771 | **0.961** | 0.660 |
| | *Infra.* | 0.829 | **0.849** | 0.726 |
| | *Indoor* | 0.872 | **0.925** | 0.796 |
| | *AVG* | 0.824 | **0.912** | 0.727 |
| **Alert %** | *Vehicle* | **100.00%** | **100.00%** | 93.33% |
| | *Infra.* | **100.00%** | **100.00%** | 96.67% |
| | *Indoor* | **100.00%** | **100.00%** | **100.00%** |
| | *AVG* | **100.00%** | **100.00%** | 96.67% |

Table 1. *Performance of TBS on all datasets. Alert % refers to the number of correctly alerted people per clip. Both the Default and Reactive mode are able to correctly detect all occurrences. Certain mode significantly increases precision, resulting in fewer false positive alerts, at a cost of Alert %. Consideration of the cost of false alarms versus missed alerts is required in real world usage.*

### 5.1 Difficult Scenarios and Failure Cases

Performance on the 'Outlier Clips' set (described in Section 4.1) is shown in Table 2 and Figure 8. As expected, the performance is significantly lower than that on the three main datasets, however the system still detects the majority of occurrences in *Default* and *Reactive*, with the majority of these alerts having a delay of <1s. This performance would still be beneficial when used to augment existing safety measures, especially considering the difficulty that a human has in perceiving a detection in these examples. Additionally, these results highlight the importance of correct installation, as situations where the camera's field of view does not line up well with the actual area of interest can result in similar cases to Figure 7.

## 6 Discussion

The TBS was able to successfully detect and alert the user to the presence of all people in the test dataset in both *Default* and *Reactive* modes. However, there is a clear trade off between alert delay and frequency of false positives that must take into account the actual usage of the system in the real world. The TBS is a human in the loop safety system that does not operate in isolation, but instead augments the existing perception of workers with regards to their ability to detect hazards in their working environment. Human reaction time to a visual stimulus in perfect conditions has been shown to be between 200-250ms [17, 18]. This time grows significantly in the presence of distractions, with the addition of just two coloured images alongside the target image of a stop sign increasing reaction time to over 550ms [19]. In the presence of tasks requiring significant mental focus, such as those carried out in all heavy industries, workers can even experience inattentional blindness, resulting in the complete missing of hazards altogether [20]. The *Alert Delays* reported in this work are comparable to those of a human applying their entire focus on the task of detecting a hazard without any distractions. The results from this work are taken from real world use cases in which the system is currently being applied, and involve cluttered and distracting environments in which human reaction time has been shown to greatly deteriorate. The TBS can achieve these results whilst not suffering from fatigue of lapses of attention.

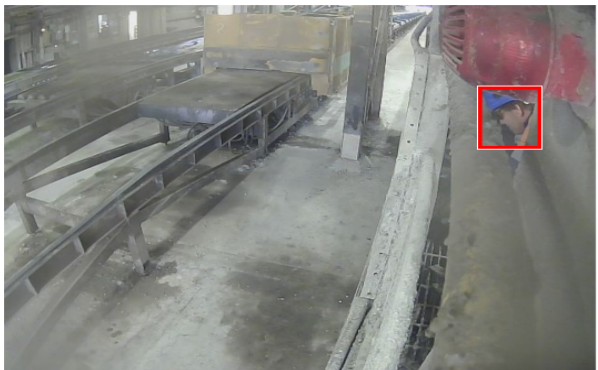

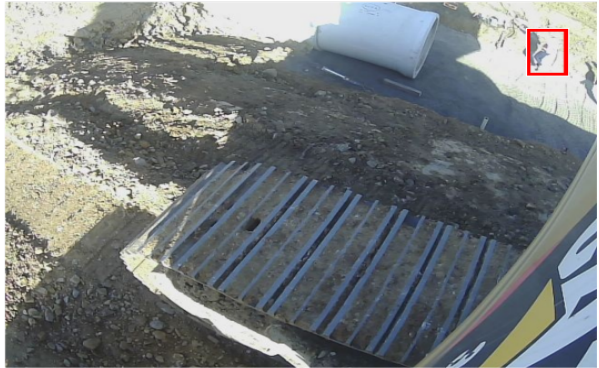

Figure 7. *Example frames from videos in 'Outlier Clips' set with the person highlighted in each frame. Both (a) and (b) are the frame selected by a human labeller as being the easiest frame in each respective clip to identify the person.*

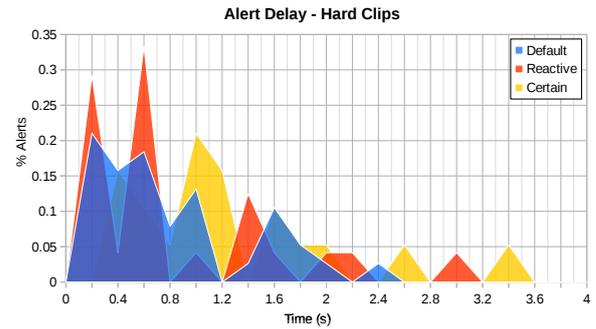

Figure 8. *Delays when testing on the 'Outlier Clips' data subset, displaying noisier alerts as expected.*

| Outlier Clips | Mode | | |
|---|---|---|---|
| | *Default* | *Reactive* | *Certain* |
| **Precision** | **0.915** | 0.771 | 0.884 |
| **Recall** | 0.463 | **0.482** | 0.378 |
| **Alert %** | **53.3%** | **53.3%** | 46.7% |

Table 2. *Performance of TBS on 'Outlier clips' (detailed in Section 4.1) which have been chosen by a human labeller as being difficult to identify any person in the video.*

An understanding of how the system is used is required when considering the importance of each metric reported in this work. Recall, a measure of how many frames were correctly classified for each person, is not as important in pratice as the metrics of alert % and delay due to how the alerts are interpreted by a human user. As each alert sent across the AlertWear network results in a two second window of notification, as described in Section 3.4, any incorrect classifications during this time are filtered out and not noticed by the user.

The results of this work additionally highlight how different modes are appropriate for different use cases. *Default* and *Reactive* both suit safety critical situations where delayed response to hazards is the crucial factor, while *Certain* would be more useful for less time pertinent uses, such as security or site access control in which false positives would want to be minimised. Whilst a number of alerts were missed in the 'Outlier Clip' set, these examples were based on videos in which a human labeller operating in perfect conditions struggled to identify the hazard, and so would likely have also been missed by a worker.

## 7 Conclusion

We have evaluated the performance of the TBS system for the detection of people in safety critical environments, validating its use as a tool for improving situational awareness in heavy industries. Evaluating the TBS system as 'fit for purpose' cannot be achieved by simply comparing Detection % on the test datasets to a given acceptable threshold. Instead, as the system is intended to augment a human's perception in real world use, it should be evaluated based on it's ability to improve this perception. It is clear when we compare the performance of TBS to human workers operating in similarly demanding environments, as discussed in Section 6, that the TBS provides a significant benefit with regards to the detection and communication of hazards in safety critical environments, and is able to do so without being subject to issues of fatigue and attention.

With the benchmark set in this paper, future work in testing the complete TBS will involve evaluation of additional detection classes and validation of additional sensing nodes, and will be conducted on a larger and more diverse dataset.

## Acknowledgments

The authors would like to thank Laing O'Rourke and the Engineering Excellence Group for their support and contributions, as well as everyone involved at Presien, except for Jason.